\documentclass{article}
\usepackage{ijcai23}

\usepackage{times}
\usepackage{soul}
\usepackage{url}
\usepackage[hidelinks]{hyperref}
\usepackage[utf8]{inputenc}
\usepackage[small]{caption}
\usepackage{graphicx}
\usepackage{amsmath}
\usepackage{amsthm}
\usepackage{booktabs}
\usepackage{algorithm}
\usepackage{algorithmic}
\usepackage[switch]{lineno}

\title{Finding Uncommon Ground: \\A Human-Centered Model for Extrospective Explanations}
\author{
Laura Spillner$^1$
\and
Nima Zargham$^1$\and
Mihai Pomarlan$^2$\and
Robert Porzel$^1$\And
Rainer Malaka$^1$
\affiliations
$^1$Digital Media Lab, TZI, University of Bremen\\
$^2$Linguistics Department, Univserity of Bremen
\emails
\{laura.spillner, zargham, pomarlan, porzel, malaka\}@uni-bremen.de
}
\begin{document}

\maketitle

%##############################
%##############################
\begin{abstract}
The need for explanations in AI has, by and large, been driven by the desire to increase the transparency of black-box machine learning models. However, such explanations, which focus on the internal mechanisms that lead to a specific output, are often unsuitable for non-experts. To facilitate a human-centered perspective on AI explanations, agents need to focus on individuals and their preferences as well as the context in which the explanations are given. This paper proposes a personalized approach to explanation, where the agent tailors the information provided to the user based on what is most likely pertinent to them. We propose a model of the agent's worldview that also serves as a personal and dynamic memory of its previous interactions with the same user, based on which the artificial agent can estimate what part of its knowledge is most likely new information to the user. 
\end{abstract}

\begin{figure}[t]
    \centering
    \includegraphics[width=1\linewidth]{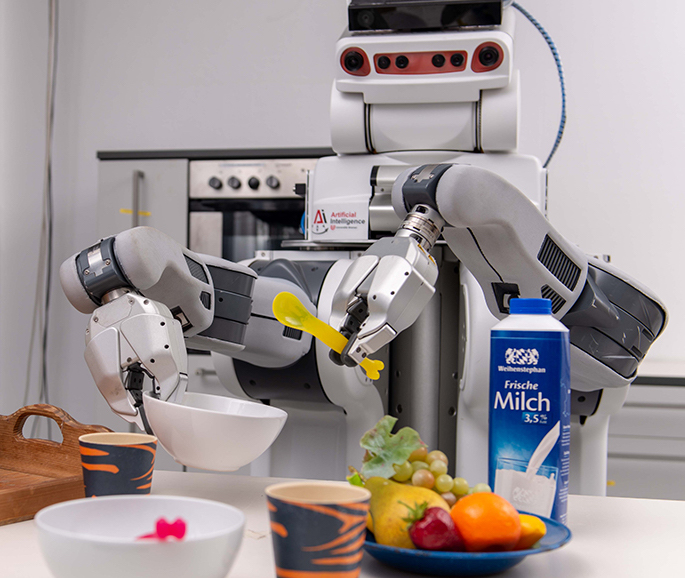}
    \caption{In domains such as household robotics, human users interact with AI agents repeatedly over time. When the two disagree, helpful and succinct explanations of the AI's behavior should be tailored to the user based on its memory of their previous interactions.}~\label{fig:teaser}
\end{figure}

%##############################
%##############################
\section{Introduction}

One major motivation of explainable artificial intelligence (XAI) is the desire to make the predictions of black-box machine learning (ML) models more transparent \cite{barredo_arrieta_explainable_2020}. Adadi and Berrada conducted a survey of the XAI literature and created an overview of common XAI methods - all of the methods presented range over the internal mechanisms of a ML model \cite{adadi_peeking_2018}, which we will refer to as {\it introspective} explanations. 

%In many studies concerning explainable artificial intelligence (XAI), the scope of the explanation ranges over the internal mechanisms that caused a particular output/behavior of a system -- comprehensive surveys have been undertaken \cite{adadi_peeking_2018,barredo_arrieta_explainable_2020}. We will refer to this type of explanation as {\it introspective}. The need for such explanations has largely been motivated by the desire to make the predictions of black-box machine learning (ML) models more transparent, with scientists or engineers of these systems often being the intended recipients of such explanations. 
%Another type that is less frequently examined concerns explanations that are given with respect to assumed expectations of another agent, e.g., a human collaborator or user of the system, that wonders why A happened instead of some tacitly expected B. This type of explanation could, therefore, be termed {\it extrospective}.
More recently, there has been an increasing need to explain AI behavior to non-expert users of AI systems. Therefore it has been proposed that XAI needs to focus more on end-users as the recipients of the explanations \cite{xu_toward_2019,liao_questioning_2020} and on how XAI methods can be evaluated from a human-centered point of view \cite{norkute_ai_2020}. This sheds a different light on estimating the quality of an explanation and how it can be tailored adequately to the given user. 
This view is supported by the argument made by several researchers in the field that explanation should be understood as interactive conversations \cite{miller_explainable_2017,feldhus_mediators_2022}. 

The usefulness of explanations can depend on the context in which they are given, including the situation and the user, as pointed out in several recent studies \cite{conati_personalized_2022,graefe_why_2022,norkute_ai_2020}. 
To foster a human-centered approach to XAI, individuals and their preferences should be considered by appropriate user models. These can encompass, e.g., a level of education, cultural background, familiarity and interest in technology, perception of the agent, mood, and social settings \cite{ZarghamProactive}. To accommodate these factors, personalization of AI explanations might be crucial \cite{conati_personalized_2022,graefe_why_2022}. This is particularly relevant in domains where users interact with the same artificial agent over a long period, such as smart home devices, robots in household and care domains, and other personal AI applications. If users ask for explanations because they expected different AI behavior, there must be some disagreement between users and AI in their respective beliefs and knowledge, which is what we call their \emph{uncommon ground}. Explanations should allow the user to realize where the uncommon ground is.
In error cases, this can empower users to correct the AI's behavior for future interactions; in other situations, it can enable them to see that they are missing important knowledge themselves. To facilitate this, explanations by the AI should reflect its experience from previous interactions with the same user.
Personalization can improve user satisfaction~\cite{wolters2009reducing,zargham2022Personalization} by using user models created with direct or indirect user input, i.e., user entry or automated systems that adapt to user behavior~\cite{mobasher2000automatic}. Although direct user inputs can have advantages, an agent that is not properly configured may be less acceptable~\cite{braun2019your}, thus personalization through indirect adaptation might be preferential~\cite{zimmermann2005personalization}. 

We argue that it is thus necessary to consider how to generate \emph{extrospective} explanations: explanations that are given with respect not only to the system itself but also take into account what the system knows about the user's expectations and draw on experience from earlier interactions. This extrospective perspective on XAI is at the heart of our work, as we focus on domains where users interact with smart software, intelligent devices, and autonomous agents/robots repeatedly over time, e.g., smart home device [citation redacted] %~\cite{zargham2022Personalization} 
and household robotics [citation redacted]% \cite{hoffnerunderstanding}
. This, in a sense, entails that users have their `own' personalized AI. As Alizadeh et al. point out, explanations in the context of repeated interactions have not been studied as much \cite{alizadeh_user-friendly_2022}. Therefore, we seek to examine this challenging problem systematically.

%##############################
%##############################
\section{Explanations in Context}
To evaluate explanations, it makes sense first to consider \emph{why} there is a need for an explanation from the user's side. We argue that in domains such as household robotics, where non-expert users commonly interact with AI systems that they generally trust, if a user asks for an explanation in a specific situation, this will likely be because the AI's behavior or recommendation differed from the user's expectation - meaning the need for explanation is motivated by the user's \emph{surprise}. 
%Liao et al. surveyed AI practitioners and developed a taxonomy of questions that users ask of XAI systems \cite{liao_questioning_2020}. They identified several different situations and goals, ranging from users wanting to increase their confidence in their own decision to trying to debug or improve the AI system. 
The XAI literature commonly differentiates between global and local explanations:
Global explanations aim to make the overall workings of a (black-box) AI model transparent \cite{adadi_peeking_2018,barredo_arrieta_explainable_2020}, e.g., through global feature importance or approximation. Here the question is about ``how'' the model works \cite{liao_questioning_2020}. Local explanations try to explain a specific instance of model prediction or agent behavior \cite{adadi_peeking_2018,barredo_arrieta_explainable_2020}, e.g., through local feature importance or salience. Here the question is about ``why'' the model arrived at this prediction \cite{liao_questioning_2020}. 
Global explanations are independent of any particular prediction. Regarding local explanations, we can further categorize explanation requests based on whether or not the user expected a different AI prediction. When users ask for a local explanation for one specific prediction, often the reason is that they expected a different outcome in this instance. 

Liao et al. surveyed AI practitioners and developed a taxonomy of questions that users ask of XAI systems \cite{liao_questioning_2020}. They identified several situations and goals, ranging from users wanting to increase their confidence in their own decision to trying to debug or improve the AI system. 
Both global and local explanations can be useful in cases where one wants to better understand how a model works, what its prediction is based on, or make its workings transparent for ethical purposes. 
This is often the goal of researchers or developers but was also a common need by other practitioners surveyed by Liao et al.
However, as the authors also point out, for non-expert users of AI, the need for local explanations most often arises out of surprise \cite{liao_questioning_2020}. 

There are many domains in which AI is supposed to support humans in making decisions or achieving tasks, such as AI software to support medical diagnosis \cite{holzinger_what_2017}, 
but humans have to make the final choice. Even though in many such domains AI performance equals or even surpasses human performance \cite{holzinger_what_2017}, it is still often desirable for the human expert to make the final decision on whether or not to trust the AI prediction. 
If the AI prediction does not align with what the user expected, it is crucial that the AI is able to explain itself \cite{zhang_effect_2020,liao_questioning_2020}. In this case, the user's goal is to understand whether their intuition or the prediction of the system is more likely to be correct. 
In domains where users can easily recognize AI errors, e.g., everyday activity domains such as household robotics \cite{kunze2010putting}, they similarly ask for explanations in (presumed) error cases to find the source of the error. Here, the user's goal can be to adapt their behavior to the system's weaknesses or improve the AI's performance, e.g., by giving it additional information.

Suppose users ask for explanations because the AI's prediction or behavior violated their expectations. In that case, a good explanation clarifies why the AI came to a different conclusion than the user. 
The explanation given should help users in finding the source of the disconnect. In some domains, this will enable users to understand AI errors or correct agent behavior in the future. In other domains, understanding why the AI prediction did not align with their intuition might help humans to realize if and why their intuition was wrong and decide whether or not to trust the AI prediction instead.
The use case we describe essentially calls for \emph{counterfactual} explanations: instead of simply asking ``why?'', the user asks, ``why not [what I expected]?''. This is supported by Miller et al.'s argument that when humans ask for explanations, there is often an implied counterfactual that is not stated explicitly \cite{miller_explainable_2017}. 

Consider a situation in the household robotics domain where a robot has helped its human user to bake a cake for a birthday party. The robot knows that because the cake has a lot of icing, it must be stored cool until the party later in the evening. The robot finds that the cake is too big to fit in the fridge, where things that must be kept cool are usually stored, but as it is cool outside, it puts the cake (in its container) on the patio table. If the human asks the robots ``Why did you put the cake outside?'', what are they really asking? The literal answer to the question is ``the cake needs to be kept cool''. However, the human is aware of this - what they are really asking is likely ``Why did you put the cake outside, instead of where I expected it [e.g., in the fridge]?'' This is the counterfactual assumption that is asked for - the user wants to know how the situation would have to be different for the cake to be stored in the expected location (e.g., if the cake were smaller, it would fit in the fridge).

There have been a number of works recently that aim to generate counterfactual explanations in different use cases \cite{miller_explainable_2017,mertes_alterfactual_2022,tompkins_effect_2022,le_improving_2022}. Yet, these approaches often encounter a problem when a) there is more than one possible counterfactual and b) the counterfactual case, that is, the prediction or behavior expected by the user, is only implied and unknown to the AI. If the decision to be explained was a choice between two options, then a counterfactual explanation simply needs to state what would have caused the decision to flip from what was actually chosen to the alternative. However, if there are more than two options such as A, B, and C, then generating a counterfactual explanation involves choosing which alternative case should be looked at. If the user explicitly states which alternative they expected, this can be used as the basis for the counterfactual explanation. However, in many situations, humans will not provide this information explicitly but only imply it or consider it obvious (it is often obvious to humans due to context knowledge, but not to the artificial agent). We can consider several strategies to discover the counterfactual: the simplest approach is to explicitly ask users what other outcome they expected. Alternatively, it is possible to try to predict the counterfactual case, e.g., by constructing a mental model of the user's knowledge or by collecting training data from a variety of different situations. 
In this paper, we propose to tailor the explanation in order to present to the user that piece of information that is most likely to be surprising to them.

\begin{figure}[t]
    \centering
    \includegraphics[width=\linewidth]{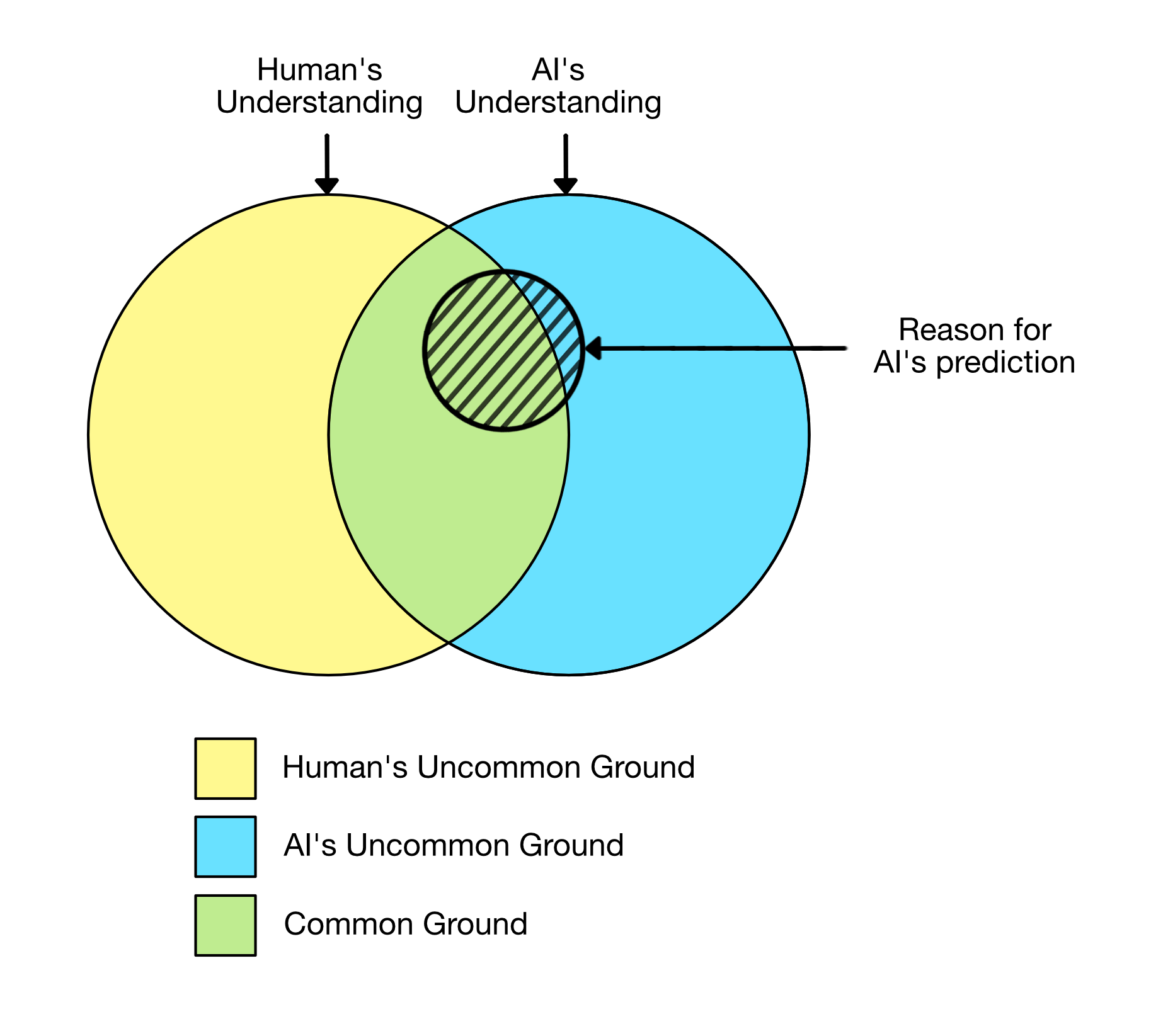}
    \caption{The left circle represents the human's understanding, that is, their knowledge and beliefs, worldview, values, etc., while the right circle represents that of the AI. Their intersection (green) is the common ground, i.e., knowledge both parties share. That part of their knowledge and understanding that is not shared by the other agent is the human's and AI's respective uncommon ground. The lined circle represents the subset of the AI's knowledge used in its reasoning to arrive at the current prediction. In the situation shown here, most of the knowledge used for the prediction is shared by the human (the part of the lined circle overlaying the green area), but the AI also reasoned with some facts or beliefs which the human does not agree with (the part of the lined circle that is in the blue). Because of this, the AI arrives at a different prediction than what the human expected.}~\label{fig:uncommon}
\end{figure}

%#############################
%######### Section ###########
%#############################
\section{Uncommon Ground}
Both the human agent and the AI or artificial agent have some understanding of the world and the current situation. In the following sections, we will consider the case in which this understanding is modeled explicitly and AI decisions are based on symbolic inference and reasoning. Although the framework we present could be applicable to ML models as well, our contribution is based on the symbolic approach. In the household robotics domain, we usually model the AI knowledge in an Abox (``assertions'': statements about concrete items in the current situation) and Tbox (``terminology'': concept definitions, general knowledge). %However, the same assumption applies to ML models or neural networks. 
In the following sections and the given example, we will use a simplified model in which the knowledge about the world, in general, contains rules about how the world works (e.g., ``if an item is a cake, then its taste is likely sweet''), while the knowledge about the specific situation contains facts about it (e.g., ``item A is a cake''). Therefore, we will assume that the agent's and the human's understanding of the world can be modeled as such facts and rules, and that both of them are able to arrive at new conclusions about the situation through logical reasoning.

Presumably, the respective world views of humans and AI mostly overlap. Yet, when the need for explanation arises from surprise on the user's side, there must be at least some small area where they do not agree: we will call this their \emph{uncommon ground}, as opposed to the common ground that is their shared knowledge. In language use, the importance of \emph{common ground}\footnote{It is important to note that \emph{common ground} is not the same as \emph{common knowledge}. While the knowledge in the common ground is shared between both human and AI, they do not necessarily know that the other shares this knowledge. In epistemic logic, common knowledge goes beyond common ground, describing knowledge that is not only known by all in the group but also known by all to be known by all. Thus, the common knowledge of the two agents is a subset of their common ground.} has been well established \cite{clark1996using}. 
In the case of AI systems using logical reasoning, when there is disagreement, this means that somewhere in the chain of reasoning of the agent, there must be a fact or a logical step with which the human would disagree (Figure \ref{fig:uncommon}). Note that for the purposes of this framework, we consider beliefs, values, moral guidelines, cultural standards, etc., to be part of this knowledge and count them as `facts' or `rules' from the the agent's point of view, even if they do not constitute (subject-independent) facts about the physical world. Thus, the uncommon ground that leads to unexpected behavior could be that one agent is not aware of some property of the current situation that the other knows, such as ``we are out of milk'', but it could also be a difference in what they consider polite, good, or proper, such as that one of them does not agree with the `rule' that ``guests should remove their shoes before entering a home''. 
%In general, among all the knowledge or the parts of the model that influenced the decision of the AI, there must be some part of it that belongs to the uncommon ground and is not shared by the user.
Even though the agent does not know what outcome or prediction the user expected, the part of their decision-making that belongs to the uncommon ground must necessarily be the explanation for the counterfactual: if \emph{this} had been different and had aligned with the belief of the user, then the prediction of the agent would have been the same as the prediction of the user.

Consider again the example of the birthday cake introduced in the previous section. If the human knows that the cake should be put in a cool place, but they expected the robot to put it in the fridge, then they were unaware that the cake is too big to fit into the fridge. This is the uncommon ground in this scenario: the robot knows a rule (items of this size do not fit in the fridge) that is not part of the human's knowledge. In this case, this is not because the human disagrees with this rule but simply because they are unaware of it. We can think of other examples where instead, the human disagrees with something the robot believes: maybe they are aware that the cake is too big for the fridge, but they don't consider the outside an appropriate place to store a cake. In this case, they would likely want to correct the robot's behavior for the future. Finally, a third situation would be one where the human is unaware of a certain property of the current situation: perhaps they do not know that outside on the patio, it is quite a bit cooler than inside because the previous days were much warmer. The robot knows a fact (the temperature on the patio) that the human has not perceived yet. This poses the question: How can the robot guess which situation is more likely, and where the uncommon ground is?

\begin{figure}[t]
    \centering
    \includegraphics[width=1\linewidth]{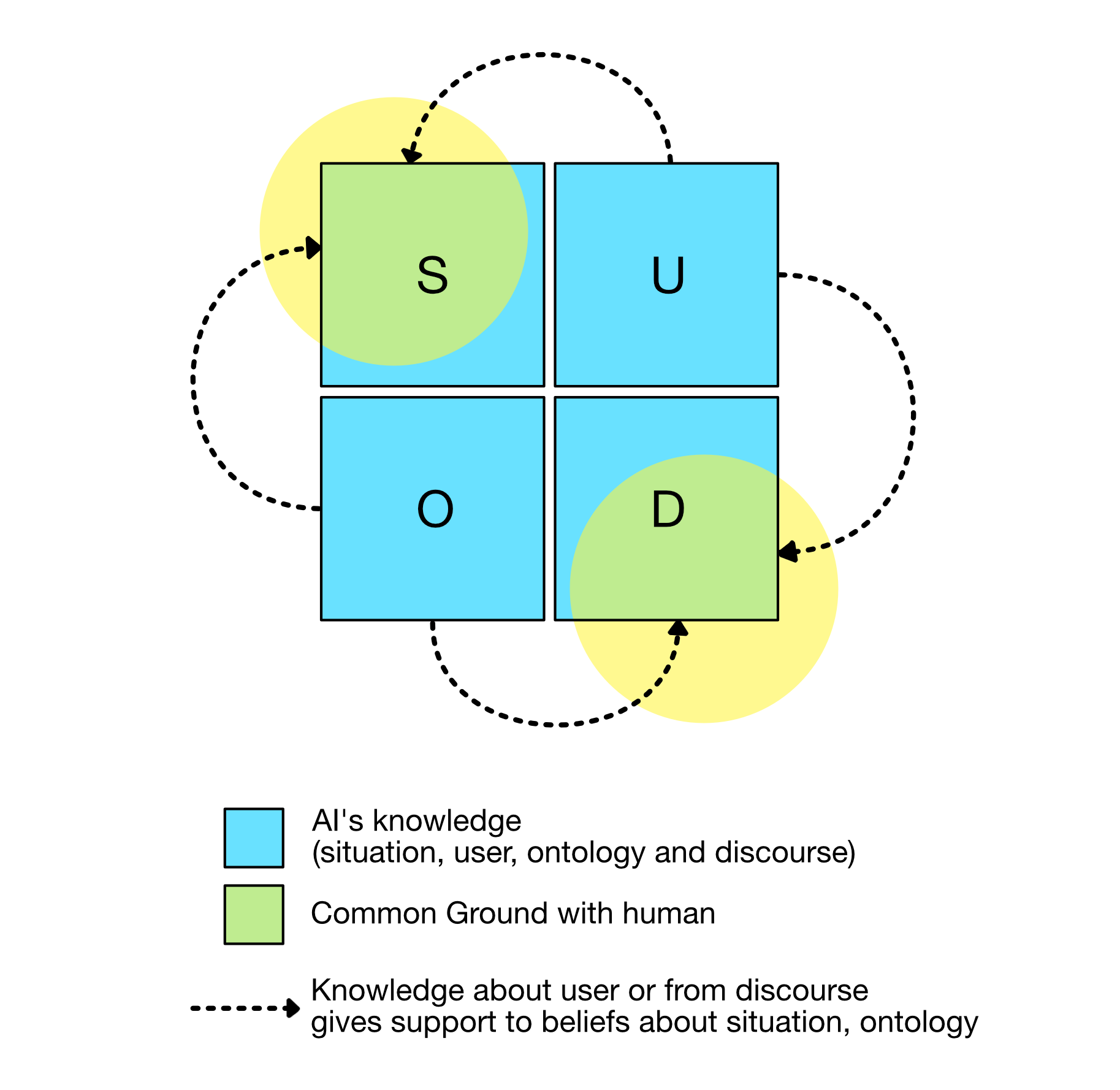}
    \caption{The internal model of the agent's knowledge is divided into \textbf{situational context} (S), \textbf{user context} (U), \textbf{discourse context} (D), and \textbf{ontological context} (O). The knowledge and beliefs of the user also include a situational context and an ontological context (yellow), which overlap with those of the agent, but are unknown to the agent. Their respective intersections represent what the agent and the user agree on. Somewhere in the set difference (the uncommon ground) must be the reason for the user's surprise. Information from both user context and discourse context can give support (dotted arrows) to facts stored in the situational and ontological contexts, allowing the agent to reason that these facts are less likely to be part of the uncommon ground.}~\label{fig:sudo}
\end{figure}

We propose a model of the agent's worldview that also serves as a personal and dynamic memory of its previous interactions with the same user. Based on this memory, the agent can judge what part of its knowledge is more likely to be shared by the user. 
We base our model on the SUDO model proposed by Porzel, in which there are four areas of context (Figure \ref{fig:sudo}) \cite{porzel2010contextual}: 

\begin{itemize}
	\item \textbf{Situational Context}: Includes knowledge specific to the current situation, such as the facts observed by the agent. In knowledge engineering, this is termed the Abox and represents instances.
    %includes facts about specific instances of items.% instead of general knowledge facts about the world.
    \item \textbf{User Context}: Includes the knowledge about the user, their physical or cognitive abilities (e.g., young children or elderly users), current activity, physical or emotional state (e.g., stress, sadness, fatigue), their personality or preferences, (e.g., on privacy, agency, etc.) as well as other humans (e.g., family, guests).
	\item \textbf{Discourse Context}: includes information on previous interactions with the user: their requests, the AI's predictions or actions, and the user's behavior and reactions.
    \item \textbf{Ontological Context}: In contrast to the situational context, this includes the general understanding of the world, such as commonsense knowledge or rules that can be used to reason with; also known as Tbox. 
\end{itemize}

%there are: the facts the agent knows, the facts the user knows, and probably also the actual facts about the world %(e.g. something happening in the world due to actual facts might very well surprise both agent and user... but lets ignore this for now)

The reason for the uncommon ground might be that either agent or user is missing some knowledge, or have specific beliefs that the other does not share (which may or may not be wrong, or might be a matter of perspective such as social etiquette).
Many works in XAI focus on explanations in the context of errors of the AI, e.g., Alizadeh and Pins developed a taxonomy of different situations in which users ask for an explanation based on why an intelligent assistant made a mistake \cite{alizadeh_user-friendly_2022}. However, if the disconnect between agent and user is based on the uncommon ground in the situational context, this does not necessitate an error by the AI: either agent or user might be basing their decision on outdated knowledge; %of the situation in cases where only one of them has observed a new fact; 
or there might be a property of the situation at hand that is not available to or perceivable by the AI. Similarly, a disagreement in the ontological context does not necessarily need to be a bug, but instead might be because of a cultural preference by the user (e.g. ``guests should keep their shoes on in my house'') that does not align with the default programming of the AI, and which is not shared by many other users.

\begin{figure*}[t]
    \centering
    \includegraphics[width=1\linewidth]{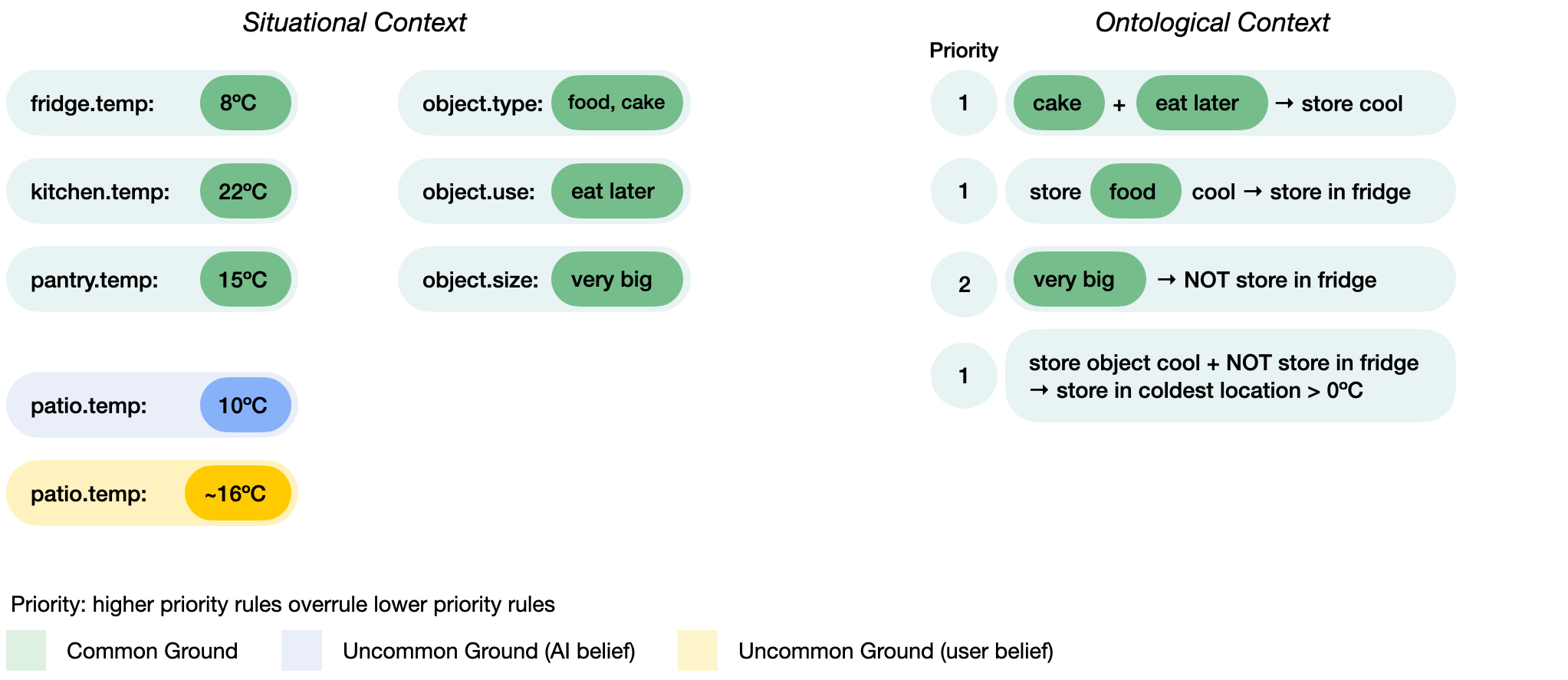}
    \caption{A simple representation of what (a subset of) the situational and ontological contexts of the AI's and the human's knowledge in the birthday cake example (see section 2) could look like. The situational context contains facts that the agents believe about the world's current state, including the properties of objects or locations. The ontological context includes rules describing how the agents believe that the world works and how they should act in certain situations; e.g., when a cake has to be saved for later, it should be stored in a cool location and not left out in a warm room. In this model, rules can contradict and outrank each other. A rule with priority level 2 precedes one with priority level 1. In this case, the rule that very big items cannot be stored in the fridge outranks the rule that food items should be put in the fridge to store cool. Here, the only difference in their respective worldview is that the AI agent is aware that it is quite cool outside today.}~\label{fig:example_context}
\end{figure*}

\begin{figure*}[t]
    \centering
    \includegraphics[width=1\linewidth]{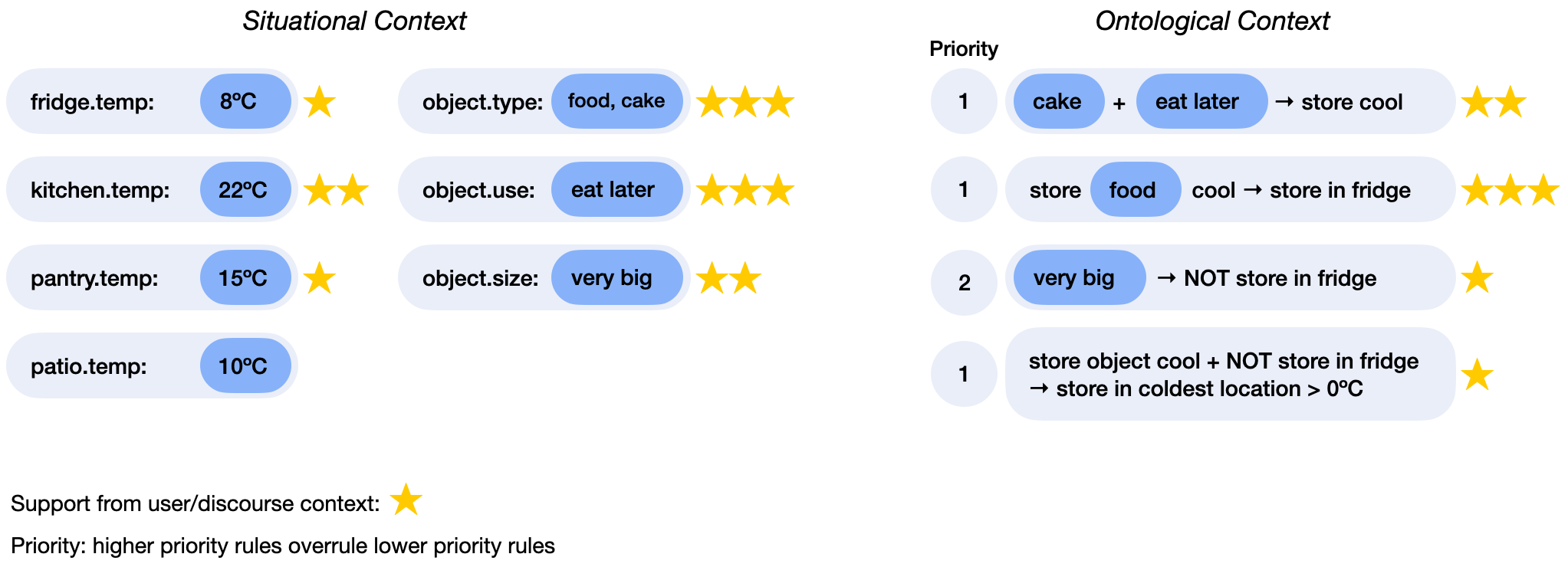}
    \caption{The same example situation as shown in \ref{fig:example_context}, but from the point of view of the AI agent. The AI does not know what the human believes or where the uncommon ground is. The stars indicate that a piece of knowledge has received support from what the agent knows about the user or from previous interactions with the user. In this case, three stars in the situational context indicate that a fact has been communicated to the robot by the user, two stars indicate that the robot knows that the user has perceived this fact as well, and one star indicates that this is a common fact which the user has agreed with in past interactions. In the ontological context, three stars would be a rule that the user had explicitly supported, while two stars indicate a rule that the agent knows the user has employed themselves in the past, and one star indicates that the agent had successfully used the rule in the past when the user did not disagree with its actions.}~\label{fig:example_support}
\end{figure*}

In the ontological context, we considered three layers: firstly, the AI will have some fixed knowledge that cannot be edited by a user, e.g., basic rules to prevent unsafe agent behavior. Other rules, forming the second layer, might be open to editing: social customs or common wisdom might be ``pre-installed'' on a household robot, but if it conflicts with the user's beliefs, could be changed based on their preferences. An example of this is a household robot by a western company that believes that meals are eaten sitting on chairs around the table, but this rule could be changed by users from other cultural backgrounds who prefer to eat sitting around a table on the floor. %We, therefore, consider this to be the second layer of knowledge. 
On the third layer, users have the option to teach an intelligent device completely new knowledge, as long as it does not conflict with any previous beliefs but was simply not available to the agent at all. 

Figure \ref{fig:example_context} illustrates what the respective sets of knowledge of both a human and an AI agent in the birthday cake example might look like. Of course, this is only a small subset of what either agent knows about the world and the situation at hand, we focused here on those facts and rules which are directly relevant to the situation and did not consider different levels. In our example, the agent knows or rather believes a number of facts about the object at hand (the birthday cake) as well as the possible storage locations (fridge, different rooms, patio). The rules can be understood to illustrate the reasoning process of the AI in this situation when read from top to bottom. Let us consider a situation in which the human agrees with almost all of the rules and facts that the AI believes, but is not currently aware of the fact that the outside temperature is quite cool. If they were made aware, they would come to the conclusion that the robot is in fact right to put the cake on the patio. Instead, they had expected it to place the object in the hallway, which they thought was just as cool as the outside. In this case, the uncommon ground was due to incorrect knowledge on the side of the user, and not due to AI error. 

In our framework, the situational and ontological contexts are influenced by two further contexts that describe how the agent understands their relation to the user as another intelligent agent.
The user context and discourse context impact the knowledge in the situational and ontological context: When the user corrects the agent's beliefs, makes it aware of new facts, or adds their own preferred rules, the knowledge that the agent has is changed based on their interaction. Facts asserted and rules edited by a user become common knowledge between the agent and user, meaning that the agent knows that the user shares this knowledge and can, in the future, assume that this is not part of the uncommon ground.
Moreover, the information from the user context and the discourse context can provide \emph{support} for the knowledge in the ontological or situational context: If some facts or rules are used in successful interactions, then the knowledge that the AI used to arrive at this decision is likely knowledge that is shared by the user. 
Therefore, we propose that the agent's internal model should be able to represent this kind of context-based support of existing knowledge (see Figure \ref{fig:sudo}), e.g., by adding weight to existing connections. In Figure \ref{fig:example_support}, we illustrate the same knowledge space as in Figure \ref{fig:example_context}, but from the point of view of the AI. The AI agent does not know which part of its knowledge is shared with the user, where the uncommon ground is, or what the human believes. However, their beliefs have gained support from previous interactions with the user as well as their knowledge of the user's previous actions and statements. The support is illustrated by stars on the respective rules and facts. To keep the graphic simple, we do not distinguish between user and discourse context. For this example, we imagine that all rules and facts have received some support except the one that describes the temperature outside (as the user is wrong about this fact, it cannot have received any support by interactions with the user). The other facts and rules have received varying levels of support, depending on whether or not the information was explicitly communicated by the human (e.g., the robot knows for a fact that the human knows that the item is a cake that is to be eaten later), is based on observations (e.g., the human has been in the kitchen. Therefore the robot can assume that the human knows the temperature there), or is based on previous successful interactions (e.g., the robot had previously put food items in a different location when they did not fit in the fridge, in that case, there was no problem). 

%%%%%%%%%%%%%%%%%%%%%%%%%%%%%%
%%%%%%%%%% SECTION %%%%%%%%%%%
%%%%%%%%%%%%%%%%%%%%%%%%%%%%%%
\section{Generating Extrospective Explanations}
Using the approach of modeling shared beliefs, we differentiate between two kinds of explanation-generating processes: Typically, when AI models or intelligent agents are asked to explain their predictions or behavior, generating this explanation is an introspective process. By that, we understand that the explanation is based on the AI's internal model, set of beliefs, or knowledge base. 
However, if an artificial agent also considers what it knows about the user's (another intelligent agent) beliefs, this becomes an \emph{extrospective} process.
 
When generating local explanations for end-users, we can assume that it is usually not possible to present an explanation of everything that influenced the decision or the entire chain of reasoning of the AI. Human-centered explanations ought to be short and relevant and not include superfluous or obvious information \cite{miller_explainable_2017,miller_explanation_2018,feldhus_mediators_2022,liao_questioning_2020}. When choosing what information is relevant, the standard (introspective) approach is to choose those aspects that have the most significant influence on the prediction, e.g., the most salient part of the input features \cite{abdul_trends_2018,adadi_peeking_2018,samek_explainable_2019,gilpin_explaining_2019}. When working with agents that reason logically, one might choose the last logical step in the reasoning process or the most specific rules\footnote{We employ defeasible logic, which is a kind of non-monotonic reasoning. Therefore, rules can have priority over other rules if they constitute a more specific exception to a more general rule, e.g., a general rule is ``birds can fly'' but it is outranked by the exception ``penguins are birds but cannot fly''.}. 

In the birthday cake example, there are several introspective ways to choose which piece of knowledge or belief should be presented to the user if they ask why the robot put the cake on the patio. One reasonable option is to present the fact about the situation that had the most significant influence on the decision, e.g., by looking at which statements were used most often or the last step in the reasoning process. We can present the last step in the robot's reasoning process since that is closest to the final decision: ``as the cake needs to be stored cool and cannot be stored in the fridge, I stored it in an alternative location''. However, this would still not inform the user of the temperature outside. We also considered presenting as an explanation the most specific rule, that is, the one that is highest in priority and constitutes an exception to other rules. In the situation presented in Figure \ref{fig:example_context}, this would be the rule that items of the cake's size do not fit in the fridge. In an alternative scenario where the user is aware of the temperature but does not know that the cake is too big to fit in the fridge, this would likely be the appropriate piece of information, but in the situation presented, it is not helpful. In this specific scenario, presenting a fact that strongly influenced the decision (the temperature outside) would indeed be helpful to the user. However, the same fact would not have been helpful to another user who is aware of the temperature but ignorant of the size problem. The reason lies in the fact that it depends entirely on the user which piece of knowledge used by the robot is in the uncommon ground. 

Suppose that the goal in explaining surprising AI predictions should be to present that part of the agent's knowledge or reasoning that is part of the uncommon ground. Then, the most salient input features or last-used rules are quite likely shared by the user. This does not help the user discover why the agent acted contrary to their expectation. If the knowledge presented in the explanation is already known to the user, then the user has learned that the agent shares some of their beliefs. However, from their perspective, it should still have arrived at the same conclusion as the user did with this set of beliefs. If, instead, the explanation includes some knowledge or beliefs the user was unaware of or did not share, then the user knows where the uncommon ground is and why the agent arrived at a different conclusion, enabling them to make informed decisions, adapt their behavior or teach the AI new knowledge. We argue that this problem can only be solved by explicitly modeling what the AI agent knows about the user and their previous interactions, which constitute the user and discourse contexts, to provide the agent with a personal and dynamic memory of their interactions. We propose that knowledge from these user-specific contexts can support knowledge in the situational and ontological contexts. If the user is unaware of the temperature outside, then this fact in the situational context does not have any support, as illustrated in Figure \ref{fig:example_support}. On the other hand, if the user does not believe in the size problem because they think that an item of this size should fit in the fridge, then this rule in the agent's ontological context would not have any support. Of course, there will not always be one definite answer to the uncommon ground problem based on the support SUDO model, there can easily be situations where several facts or rules have no support. Similarly, if the user changes their mind, facts that received some support in the past might still be part of the uncommon ground now. Finally, there is not necessarily only one piece of information in the uncommon ground, there can easily be several points of disagreement. However, we propose that our approach presents a viable first step at presenting a helpful explanation tailored to the user, which will give them the necessary information in many circumstances.

At this point, the agent still does not know what the user knows or believes or where precisely the uncommon ground was in this situation: the user has to explicitly instruct or teach the agent for it to gain that knowledge. In the future, users can update the agent's knowledge base. If they realize that they themselves were missing some correct information, they can update their beliefs and trust the agent's prediction; or if the difference is not easily resolvable, they might choose not to give the agent this kind of task in the future.
Therefore, the generation of the explanation needs to be an extrospective process: it needs to consider what the agent has learned about the user's beliefs and preferences throughout their previous interactions and then present in the explanation that piece of information that has the lowest support by the user and discourse context. The explanation could include assumptions by the agent that are not explicitly supported by anything it knows about the user, facts that (as far as the agent knows) are not known by the user, or rules that the agent believes but that has never before been used in successful interactions.  
%(e.g., it is not supported by anything the agent knows about the user, the user has never stated it, it has not been used in previous successful interactions, etc.).

Accordingly, extrospective explanations should be evaluated together with prospective users in different contextual scenarios. The goal of good explanations (in situations of surprise) should be to enable the user to understand how the agent's worldview differs from their own without requiring repeated follow-up questions or presenting superfluous (``obvious'' ) information to the user. In situations where the user themselves was incorrect or was missing information, they should be able to quickly learn this information from the AI when they ask for an explanation. In personalized AI domains, realizing where the uncommon ground is should empower the user to update the AI's knowledge base to align with their own preferences if they desire to do so.

%%%%%%%%%%%%%%%%%%%%%%%%%%%%%%
%%%%%%%%%% SECTION %%%%%%%%%%%
%%%%%%%%%%%%%%%%%%%%%%%%%%%%%%
\section{Conclusion}
There is a growing awareness in the XAI community that explanations ought to be considered and evaluated in the context of the user for whom they are intended. Previous works have pointed out the problem of implied-but-not-stated contrastive/counterfactual cases when humans ask for explanations \cite{miller_explainable_2017} as well as that there are many situations in which the user's goal in asking for an explanation relies on understanding why the AI prediction differed from their expectation \cite{liao_questioning_2020}.

We claim that in the domain of AI in the home (e.g., smart home devices, household robots), extrospective explanations can provide more helpful information for end-users in a more human-centered way, enabling them to understand the reason for unexpected agent behavior. We aim to achieve this by focusing on explanations for transparent, reasoning-based AI systems and explicitly modeling user and discourse context in the agent's knowledge. If the AI predicts the same outcome based on the same situational knowledge and ontology in two equivalent situations, the respective explanations could differ based on the user asking for them: For each user, the AI would output only a part of their reasoning. Specifically, the AI should output those facts or inferences that it expects are most likely to surprise the user. This should be whatever fact or logical inference the AI has the least reason to believe that the user agrees with or is aware of. 
In other domains, however, there might be different reasons why explanations should be purely introspective. There are many areas of life where AI is or will be used, in which it will be important that explanations are consistent across different users or that they are a complete representation of the AI reasoning (e.g., medical or legal problems). Therefore, it will be essential to distinguish between explanations aimed at helping individual users understand local, surprising AI predictions and explanations aimed at increasing ML systems' transparency.

%Whether or not extrospective explanations make sense depends on the context. Purely introspective explanations would always be the same for a particular prediction, which might be helpful in some domains (legal, medical, etc.). Clearly, one cannot look at the history if only one interaction exists for each user.
%Taking user and discourse support into account makes sense in domains where repeated interactions are built in by design, such as smart home devices or personal AI.  

%\appendix

%\section*{Ethical Statement}

%There are no ethical issues.

%\section*{Acknowledgments}

\bibliographystyle{named}
\bibliography{ijcai23}

\end{document}